\pdfoutput=1
\documentclass[10pt]{article}
\usepackage[margin=0.88in]{geometry}
\usepackage[T1]{fontenc}
\usepackage{mathpazo,microtype,amsmath,amssymb,graphicx,booktabs,array}
\usepackage[font=small,labelfont=bf]{caption}
\usepackage{xcolor,placeins,enumitem}
\usepackage[colorlinks=true,allcolors=blue!55!black]{hyperref}
\usepackage[numbers,sort&compress]{natbib}
\hypersetup{pdftitle={Humanized but Still True? Auditing Facts and Detection in AI-Text Humanizers},pdfauthor={Ziyuan Wang, Genliang Zhu, Xiaobo Chen},pdfkeywords={AI text humanizers, factual fidelity, machine-generated text detection, decision models, Binoculars, Fast-DetectGPT}}
\title{Humanized but Still True?\\Auditing Facts and Detection in AI-Text Humanizers}
\author{%
Ziyuan Wang\textsuperscript{1}, Genliang Zhu\textsuperscript{2}, Xiaobo Chen\textsuperscript{3}\\[0.5em]
\small \textsuperscript{1}Independent Researcher\\
\small \textsuperscript{2}Georgia Institute of Technology\\
\small \textsuperscript{3}University of Massachusetts Amherst\\[0.4em]
\footnotesize \textsuperscript{1}\href{mailto:ziyuan.wang.engineer@gmail.com}{\nolinkurl{ziyuan.wang.engineer@gmail.com}}\\
\footnotesize \textsuperscript{2}\href{mailto:sebby@gatech.edu}{\nolinkurl{sebby@gatech.edu}}\\
\footnotesize \textsuperscript{3}\href{mailto:xchen@umass.edu}{\nolinkurl{xchen@umass.edu}}}
\date{}

\begin{document}
\maketitle
\vspace{-2.6em}
\begin{abstract}
AI ``humanizers'' rewrite machine-generated text so that AI detectors classify it as human-written. Public leaderboards score these tools on detector evasion and on an embedding-similarity measure of meaning, but not on whether specific checkable details survive. We compare each source text with its rewrite and check numbers, dates, person and organization names, and spans in quotation marks, using a rule-based checker validated by two AI annotators (precision 0.84--1.00, recall 0.93--0.95; no human annotation). In the October 2026 cycle of a public humanizer leaderboard, all 11 commercial tools changed or dropped at least one of these units in 35--65\% of the documents that contained one. Most changes were to quoted spans, which in these sources were often sample phrasings, terms or rhetorical questions rather than reported speech; counting numbers and names alone, 10 of the 11 tools still altered at least one in 20\% or more of the documents. The September 2026 cycle was similar (19--50\%). An open 12B humanizer kept at least 97\% of numbers and names in fact-dense U.S. federal agency text, but reworded 67--78\% of the quoted spans, mostly attributed statements, against its own prompt. Unit retention and detector evasion were not clearly related across tools (Spearman $\rho=-0.46$, 95\% CI $-0.65$ to $0.08$), and the four released files of the open model, two of them low-bit files refined by quantization-aware training, did not show the pre-registered monotone trade-off. On federal texts where human originals, AI-styled rewrites and humanized versions cover the same content, a general-purpose decision model, Jev, asked a single zero-shot question, separated human originals from humanized rewrites with AUROC 0.864 (0.827--0.902), 0.18--0.22 higher than Binoculars, Fast-DetectGPT and the adversarially trained RADAR, while calling 24\% of the human originals AI at a 0.5 threshold. On the AI-styled rewrites by 2026 models, both perplexity-based detectors scored below chance (AUROC 0.37 and 0.39), also with line breaks removed. Hypotheses, data rules and analyses were frozen before the main runs; two earlier data contacts are disclosed. We release the checker and the per-item results.
\end{abstract}

\section{Introduction}
Paraphrasing defeats most AI-text detectors \citep{krishna2023paraphrasing,sadasivan2023reliably}, and a commercial category of ``AI humanizers'' sells exactly this service. Monthly leaderboards now run 11--14 such tools on fixed source texts and publish the outputs together with scores from 5--7 commercial detectors \citep{aihb2026,humanizerbench2026}. These leaderboards measure how often a rewrite evades detection and how close it stays to the source in embedding space. Embedding similarity, however, is insensitive to the details that matter when a humanized text is published: a changed number, a dropped name or a reworded quotation can leave the overall meaning nearly unchanged and still make the text false. The only prior study of humanizer faithfulness assessed meaning qualitatively \citep{masrour2025damage}. Studies of the trade-off between preservation and evasion measured preservation with an LLM-assigned score \citep{xu2026basemodels,cheng2025adversarial}.

On the detection side, general-purpose models have been tried as zero-shot detectors before. ChatGPT, asked whether a passage was AI-written, recognised human text well but AI text poorly \citep{bhattacharjee2023fighting}, and the implicit reward of an instruction-tuned model over its base model separates the two better than Binoculars \citep{liu2025irm}. Decision models now return calibrated choices between options, and a single generic question to one of them, Jev, already detects several kinds of alignment failure zero-shot \citep{almeida2026jev,guo2026justaskjev}. Whether such a model can tell human writing from AI writing, before and after humanization and against matched human controls, has not been tested.

We measure both sides with protocols frozen before the main runs (Section~\ref{sec:prereg}):
\begin{itemize}[leftmargin=1.3em,itemsep=1pt,topsep=2pt]
\item \textbf{Fact fidelity.} A program extracts specified units (numbers and dates, person and organization names, and spans in quotation marks) from each source text and tests whether each unit survives in the rewrite. We apply it to all 363 commercial outputs of the October 2026 leaderboard cycle and to 368 rewrites from four quantizations of an open 12B humanizer on fact-dense U.S. federal agency text.
\item \textbf{Fidelity versus evasion.} We test whether tools that evade detectors better retain fewer of these units, across commercial tools and across the four released quantizations of one model.
\item \textbf{Detection with matched human controls.} On federal texts where human originals, AI-styled rewrites and humanized versions cover the same content, we compare Jev with Binoculars, Fast-DetectGPT, a RoBERTa detector and, added after review, the adversarially trained RADAR.
\end{itemize}
Every commercial tool altered specified units in a large share of documents, mostly by rewording quoted spans; numbers and names alone still crossed our 20\% threshold for 10 of the 11 tools. We found no reliable fidelity--evasion trade-off. Of the five detectors we ran (two perplexity-based zero-shot methods, a GPT-2-era classifier, the adversarially trained RADAR and the decision model), only the decision model separated human text from AI-styled rewrites by current models with AUROC above 0.8.

\section{Related work}
\paragraph{Detectors and paraphrase attacks.}
Zero-shot detectors use statistics of a scoring language model: probability curvature in DetectGPT and Fast-DetectGPT \citep{mitchell2023detectgpt,bao2024fastdetectgpt}, and the ratio of perplexity to cross-perplexity in Binoculars \citep{hans2024binoculars}. Supervised detectors range from the RoBERTa classifier released with GPT-2 \citep{solaiman2019release} to adversarially trained models \citep{hu2023radar,mady2026conpara}. Paraphrasing evades many of them \citep{krishna2023paraphrasing,sadasivan2023reliably}, and RAID includes paraphrase and character-level attacks \citep{dugan2024raid}. DeBERTa-ConPara lists humanization tools as untested future work. It also argues that thresholds tuned on the test domain are unrealistic \citep{mady2026conpara}. Detection benchmarks can also be separable for reasons unrelated to authorship: Voznyuk and Grabovoy show that line-break differences alone separate human and machine text in several scientific subsets \citep{voznyuk2026artifact}. On matched variants of the same source texts, ARB shows that Fast-DetectGPT and Binoculars detect 91--94\% of directly generated text at 1\% false positives but only 15--31\% of human text that an LLM has rewritten \citep{perrone2026arb}. Detector-guided paraphrasing evades even robust detectors, whereas plain paraphrasing can make text easier for Fast-DetectGPT to flag \citep{cheng2025adversarial}.

\paragraph{Humanizers.}
DAMAGE defines humanizers as a category, evaluates 19 tools and assesses their faithfulness qualitatively before training a robust detector \citep{masrour2025damage}. Xu et al.\ show that text from base models looks human to commercial detectors and build an iterative paraphraser. They, like the adversarial-paraphrasing attack of Cheng et al., frame the problem as a trade-off between preservation (scored by an LLM) and evasion \citep{xu2026basemodels,cheng2025adversarial}. The two public leaderboards report evasion, embedding-based meaning and readability \citep{aihb2026,humanizerbench2026}. None of these measures whether specific facts survive.

\paragraph{Models as zero-shot detectors and judges.}
Bhattacharjee and Liu asked ChatGPT to label passages as AI or human: it missed most AI text but rarely misjudged human text \citep{bhattacharjee2023fighting}. IRM uses the log-probability ratio of an instruction-tuned model to its base model, the implicit reward of preference training, as a zero-shot score and beats Binoculars on DetectRL; polishing human text with an LLM was the attack that hurt it most \citep{liu2025irm}. Jev returns a probability for each option of a typed question \citep{almeida2026jev}. Guo et al.\ used one generic question to detect ten kinds of alignment failure across 44 benchmarks, with a median AUROC of 0.886 \citep{guo2026justaskjev}. None of these tested a decision model on AI-text detection, nor any zero-shot judge on humanized text with matched human controls.

\section{Setup}
\subsection{Text sets}
\textbf{Commercial outputs (F2).} The AI Humanizer Benchmark (AIHB) releases, for each monthly cycle, 33 AI-written source texts in seven categories (academic and application essays, blog posts, marketing copy, business e-mails, discussion-board posts and news articles). It also releases the outputs of 11 commercial humanizers and the human-probability scores of seven commercial detectors for every output \citep{aihb2026}. We used the September 2026 cycle for development and the October 2026 cycle (33 sources, 363 outputs, written by six source models) as the main set.

\textbf{Federal texts (F1).} For matched human controls we used U.S. federal agency prose, which is in the public domain. It has two parts. The first is all 54 modern passages of the GoHumanize test split, which pair a human original with an AI-styled rewrite by GPT-4o-mini (39 passages) or DeepSeek-Chat (15) \citep{gohumanize2026}. The second, NEW2026, consists of 38 passages from 2026 NIST news and Federal Reserve speeches and press releases. To build it we copied the GoHumanize pipeline: the same passage rule (consecutive paragraphs totalling 80--300 words) and the same ``hard'' rewriting prompt that GoHumanize used for its modern split, with the AI drafts generated alternately by GPT-6-Luna and Gemini-3.5-Flash-Lite. Unlike GoHumanize, whose 54 test passages come from 30 articles, we kept only the first passage of each article. Articles already in the GoHumanize data were excluded; the feeds yielded 38 usable articles instead of the 50 planned. The 92 passages come from 68 articles, and each has a human original and an AI draft.

The hard prompt (Appendix~\ref{app:prompts}) tells the generator to keep every fact, name, number and quotation but to write ``the way a large language model writes when asked to produce polished prose'': open most sentences with connectives such as ``Additionally'' or ``Furthermore'', turn verbs into noun phrases, prefer abstract wording, hedge, keep sentences 22--30 words long, run 20--40\% longer than the original, and use no contractions or dashes. The AI drafts are therefore AI-styled rewrites of the human originals, deliberately pushed towards stereotypical LLM prose, not free generations. They correspond to the H2L regime of ARB \citep{perrone2026arb} rather than to the usual human-versus-generated benchmark, and the detection results should be read with that in mind.

\textbf{Open humanizer.} We rewrote every F1 draft once with each of four GGUF files of \texttt{jialinyyzz/humanizer}, a 12B text-completion model: bf16, Q8\_0, Q3-QAT and IQ2\_XS-QAT \citep{humanizer12b2026}. According to the model card of the revision we used, Q8\_0 is a standard llama.cpp quantization. In the Q3 and 2-bit files, by contrast, sensitive tensors were given more bits, and the quantized model was then trained layer by layer to reproduce the bf16 model (quantization-aware training) and distilled from it. The four files therefore differ in training as well as in bit width; they are released variants, not a controlled quantization ladder. We used llama.cpp b11865 with the model's exact prompt (which states that every fact, number, name and quotation must survive unchanged) and its recommended sampling: temperature 1.0, top-$p$ 0.95, no other samplers, stopping at end-of-sequence only. All 368 rewrites ended at end-of-sequence.

\subsection{Fact checker}\label{sec:judge}
For each source text the checker extracts three kinds of unit, deduplicated by value:
\begin{itemize}[leftmargin=1.3em,itemsep=1pt,topsep=2pt]
\item numbers and dates written as digits or number words, compared by value (so \$1.9 billion = 1,900 million and 3.5 percent = 3-1/2 percent);
\item person and organization names found by spaCy \texttt{en\_core\_web\_sm}, kept if the surname, the full name, a standard acronym or a distinctive token of the name survives;
\item spans of at least three words inside quotation marks, kept only if they reappear word for word, ignoring case and punctuation. The checker does not distinguish reported speech from other quoted text, such as terms, sample phrasings or rhetorical questions.
\end{itemize}
A unit counts as altered if it is changed or missing. Dropping a hedge such as ``nearly'' does not count; rounding, a different value or a different unit does, even when the conversion is exact (``two decades'' to ``twenty years''). The rules and the code were developed only on development material and then frozen.

The checker was validated twice, both times by AI annotators; no human labels were collected. First, before any evaluation run, the AI assistant that wrote the checker listed every altered unit in 40 development pairs (20 commercial, 20 from the open model) without seeing the checker's output. Precision was 33/33 and recall 35/37, against preset thresholds of 0.90 and 0.85. Because the annotator also wrote the checker, these figures are probably optimistic. Second, after the main run, a different AI system (OpenAI Codex) audited 20 pairs drawn by a preset stratified rule: development and evaluation sets, commercial and open-model rewrites, and pairs with and without flagged units. It sealed its answers before seeing the earlier labels. It judged 26 of the 31 flagged units correct, found no unit marked as retained that had in fact changed, and found two altered units the checker had not extracted. This gives precision 0.84 and recall 0.93 against preset thresholds of 0.80 and 0.75. On the 10 evaluation pairs alone, precision was 6/8. Two of the five disputed flags are the exact unit conversions above, which the auditor counted as retained; under the frozen rule, precision is 0.90. Across all combinations of the four ambiguous rule readings, precision ranged from 0.84 to 0.90 and recall from 0.93 to 1.00.

\subsection{Detectors}
Jev (\texttt{typesafe/jev-1.13}, snapshot 20260917, through OpenRouter) received the full text as state and one choice question, ``Who produced this text?'', with two options: \emph{human}, ``written by a human author, without an AI language model generating or rewriting it'', and \emph{ai}, ``generated by an AI language model, or an AI draft rewritten by an AI tool''. We used $P(\text{ai})$ as the score. Binoculars and Fast-DetectGPT (analytic criterion) shared the official Falcon-7B / Falcon-7B-Instruct pair in bf16, and the metric functions were copied verbatim from the authors' repositories \citep{hans2024binoculars,bao2024fastdetectgpt}. The RoBERTa detector \citep{solaiman2019release} scored at most 512 tokens. After review we added RADAR \citep{hu2023radar} (\texttt{TrustSafeAI/RADAR-Vicuna-7B}), a RoBERTa-large detector trained adversarially against a paraphraser, with its official inference: at most 512 tokens and the probability of the AI class as the score. Because the official Binoculars threshold flagged almost no AI text in our development data, every threshold-based metric uses a threshold set at a 5\% false-positive rate on the 92 human F1 texts. This threshold is set on the evaluation texts themselves and is therefore an oracle operating point: it ranks the detectors at a fixed false-positive rate but does not describe deployable performance, and the main conclusions rest on AUROC.

\subsection{Pre-registration and statistics}\label{sec:prereg}
Three protocols were frozen with file hashes before the corresponding runs: feasibility, construction of the evaluation set, and analysis. Each phase had preset stop conditions. Two earlier contacts with evaluation data are disclosed. Before the hypotheses were written, an exploratory regular-expression count of digit tokens was run over both leaderboard cycles. Before the feasibility phase, a 30-text pilot of the Jev question used 10 human federal passages, 10 source texts and 10 humanized outputs from the October cycle; the thresholds for H3 were fixed before that pilot. The checker's rules and code were developed on the September cycle only, and Jev's scores on the commercial outputs are reported only descriptively. The single protocol deviation, the smaller NEW2026 set, was decided before any model was run on the federal evaluation texts. After an internal review, two descriptive analyses were added under a fourth protocol frozen before their runs: the RADAR baseline and the whitespace counterfactual for all detectors. They are marked as post hoc and do not change H1--H3. After a second review, the intervals on the federal set were recomputed by resampling articles instead of passages (below). The three primary hypotheses were:
\begin{itemize}[leftmargin=1.3em,itemsep=1pt,topsep=2pt]
\item \textbf{H1}: at least three commercial tools alter at least one number, name or quoted-span unit in $\geq 20\%$ of eligible documents;
\item \textbf{H2}: across tools, unit retention and mean commercial human-probability have Spearman $\rho\le-0.5$ with a CI that excludes 0 (H2a), and retention falls while evasion rises monotonically from bf16 to IQ2\_XS-QAT (H2b);
\item \textbf{H3}: Jev's AUROC for human originals versus 12B rewrites is $\geq 0.80$ (signal) or $\leq 0.60$ (failure).
\end{itemize}
Confidence intervals resample source texts (F2), each with all of its rewrites, or articles (F1), each with all of its passages and their versions, 2,000 times. The frozen analysis resampled F1 passages, which treats the several passages of one GoHumanize article as independent. After review we switched to articles; on the full federal set this moved no interval endpoint by more than 0.012 and changed no verdict (Appendix~\ref{app:cluster}). Proportions use Wilson intervals; for the 12B rows of Table~\ref{tab:fid}, they agree with an article bootstrap to within 0.02.

\section{Results}
\subsection{Humanizers alter specified units, above all quoted spans (H1)}
\begin{figure}[t]
\centering
\includegraphics[width=\linewidth]{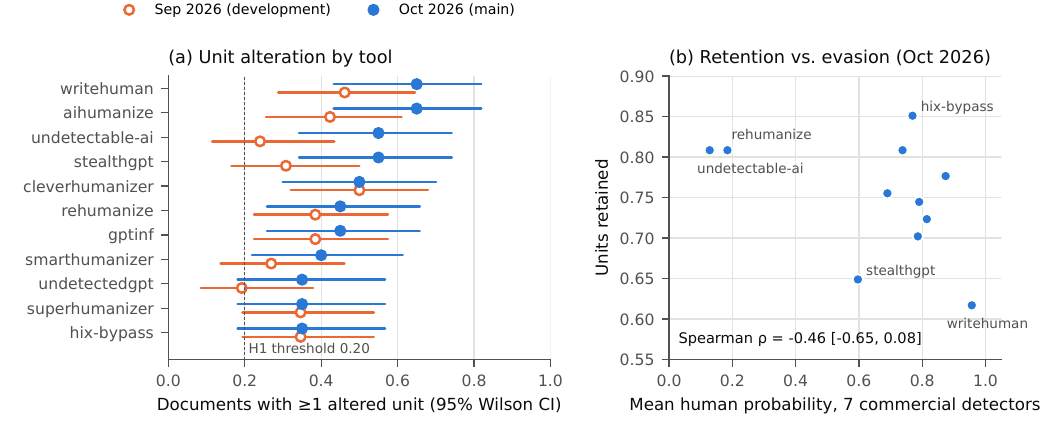}
\caption{(a) Share of documents with at least one altered unit, by tool, for the development (September) and main (October) leaderboard cycles. (b) Unit retention versus evasion in the October cycle (H2a); the CI resamples source texts.}\label{fig:fid}
\end{figure}

Every commercial tool altered at least one unit in 35--65\% of the 20 October sources that contained one (Table~\ref{tab:fid}, Figure~\ref{fig:fid}a). H1 therefore holds for all 11 tools, not just the three required, with Wilson lower bounds of at least 0.18. The September cycle used different source texts and gave 19--50\%. Quoted spans were affected most: no tool kept more than 4 of the 15 quoted-span units word for word, and 10 tools reworded at least 12 of them. Losses of numbers ranged from 2/56 to 13/56 units and losses of names from 0/23 to 8/23.

Quoted spans drive part of this result, and most of them are not reported speech. Only 6 of the 20 eligible sources contain one. Of the 15 spans, 6 are statements attributed to named people in two news articles. The other 9 are four sample lines for asking a manager for a promotion in a blog post, two placeholder achievements in a cover-letter template, a term (``discretionary time off'') and two questions an author might ask. Rewording such a span changes text that the source put in quotation marks, but it need not make the rewrite false. Counting number and name units alone, the share of documents with an alteration was 11--58\%, and 10 of the 11 tools stayed at or above 20\%; only hix-bypass fell to 11\% (Appendix~\ref{app:quotefree}). H1 therefore does not depend on quoted spans.

The open model behaved differently on fact-dense federal text. It lost at most 2.8\% of number units and 1.1\% of names, and 12--19\% of passages had any alteration (6--11\% without quoted spans; Table~\ref{tab:fid}). It still reworded 12--14 of the 18 quoted spans, contrary to its own prompt; here 13 of the 18 are statements attributed to named people, 2 are passages quoted from official documents and 3 are terms or mottos. In one development example a Q3 rewrite turned ``Champlain Towers South'' into ``Champlian Towers South''. Building names are outside our fact categories, so slips of this kind are not counted.

\begin{table}[t]
\centering\small
\caption{Unit alteration. Commercial tools: AIHB October 2026, 20 eligible sources per tool. 12B: the open humanizer on 90 eligible federal drafts; QAT marks the two low-bit files refined by quantization-aware training. Columns 4--6 give altered/total units; ``Quoted'' counts spans in quotation marks, whether or not they are reported speech. Evasion is the mean human probability over seven commercial detectors (higher means more often judged human).}\label{tab:fid}
\begin{tabular}{lrrrrrr}
\toprule
Tool & Altered docs & Rate [95\% CI] & Numbers & Names & Quoted & Evasion \\
\midrule
aihumanize & 13/20 & 0.65 [0.43, 0.82] & 11/56 & 2/23 & 15/15 & 0.79 \\
writehuman & 13/20 & 0.65 [0.43, 0.82] & 13/56 & 8/23 & 15/15 & 0.96 \\
stealthgpt & 11/20 & 0.55 [0.34, 0.74] & 13/56 & 6/23 & 14/15 & 0.60 \\
undetectable-ai & 11/20 & 0.55 [0.34, 0.74] & 5/56 & 1/23 & 12/15 & 0.13 \\
cleverhumanizer & 10/20 & 0.50 [0.30, 0.70] & 7/56 & 5/23 & 14/15 & 0.81 \\
gptinf & 9/20 & 0.45 [0.26, 0.66] & 8/56 & 1/23 & 15/15 & 0.79 \\
rehumanize & 9/20 & 0.45 [0.26, 0.66] & 4/56 & 2/23 & 12/15 & 0.18 \\
smarthumanizer & 8/20 & 0.40 [0.22, 0.61] & 2/56 & 8/23 & 13/15 & 0.69 \\
hix-bypass & 7/20 & 0.35 [0.18, 0.57] & 3/56 & 0/23 & 11/15 & 0.77 \\
superhumanizer & 7/20 & 0.35 [0.18, 0.57] & 2/56 & 2/23 & 14/15 & 0.74 \\
undetectedgpt & 7/20 & 0.35 [0.18, 0.57] & 7/56 & 1/23 & 13/15 & 0.87 \\
\midrule
12B bf16 & 11/90 & 0.12 [0.07, 0.21] & 5/247 & 1/378 & 12/18 & -- \\
12B Q8\_0 & 15/90 & 0.17 [0.10, 0.26] & 7/247 & 1/378 & 13/18 & -- \\
12B Q3-QAT & 16/90 & 0.18 [0.11, 0.27] & 7/247 & 4/378 & 12/18 & -- \\
12B IQ2\_XS-QAT & 17/90 & 0.19 [0.12, 0.28] & 7/247 & 2/378 & 14/18 & -- \\
\bottomrule
\end{tabular}

\end{table}

\subsection{No reliable fidelity--evasion trade-off (H2)}
Across tools, retention and evasion were negatively but not reliably correlated: $\rho=-0.46$, CI $-0.65$ to $0.08$ (Figure~\ref{fig:fid}b). H2a is therefore not supported. The two tools that evaded detectors least (mean human probability 0.13 and 0.19) altered facts in 55\% and 45\% of documents, so low evasion did not imply high fidelity. Within the open model, retention was 0.972, 0.967, 0.964 and 0.964 from bf16 to IQ2\_XS-QAT, and evasion (the share of rewrites below the 5\%-false-positive threshold, averaged over Jev, Binoculars, Fast-DetectGPT and RoBERTa) was 0.73, 0.73, 0.75 and 0.74. The four released files therefore did not meet the pre-registered monotone condition, and H2b is not supported. Between bf16 and IQ2\_XS-QAT, retention differed by 0.008 (CI $-0.004$ to $0.020$) and evasion by $-0.003$ ($-0.065$ to $0.048$). The data thus rule out a loss of more than about two percentage points of retention, not a smaller one; descriptively, the share of passages with any alteration rose from 0.12 to 0.19. Nor does this comparison isolate bit width: the two low-bit files were trained to reproduce the bf16 model after quantization, which according to the model card recovers much of what a standard quantization of the same size loses. Standard Q3 or 2-bit files may behave differently.

\subsection{A decision model detects humanized text that four other detectors miss (H3)}
\begin{figure}[t]
\centering
\includegraphics[width=\linewidth]{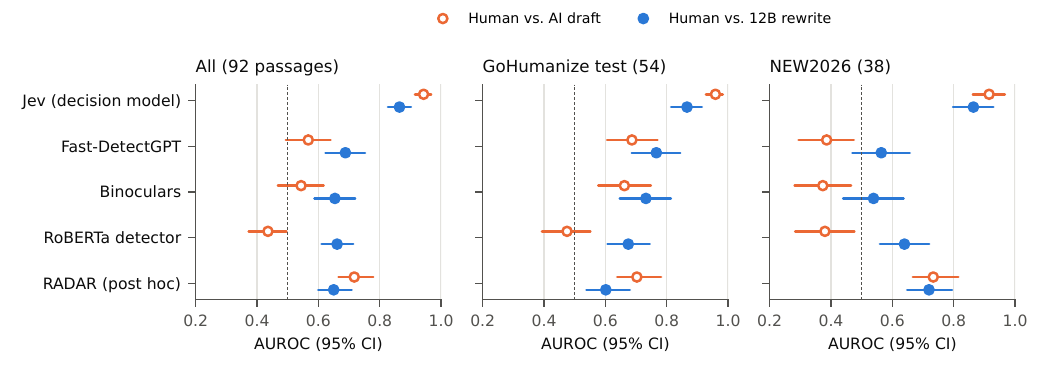}
\caption{AUROC of human originals versus AI drafts (hollow) and versus 12B rewrites (filled), for all federal passages and for each source. The dashed line is chance. Intervals resample articles; the per-source intervals are descriptive. RADAR was added after review.}\label{fig:det}
\end{figure}

On the matched federal texts, Jev separated human originals from 12B rewrites with AUROC 0.864 (0.827--0.902), which supports H3 (Table~\ref{tab:det}). At a 5\% false-positive rate it flagged 57\% of rewrites and 73\% of AI drafts. The other four detectors reached 0.65--0.69 on rewrites and flagged 5--20\% of them. Jev's AUROC was higher than that of Binoculars by 0.211 (0.132--0.294), than that of Fast-DetectGPT by 0.176 (0.099--0.256) and, in the post-hoc comparison, than that of RADAR by 0.215 (0.157--0.264). RADAR separated the AI drafts better than the perplexity detectors did (0.717, 0.666--0.779), but humanizing lowered its AUROC, as it lowered Jev's. Figure~\ref{fig:det} shows the results separately for each source of the federal set. At the natural 0.5 threshold Jev called 24\% of the human originals AI (median $P(\text{ai})$ 0.37), against 91\% of the drafts and 80\% of the rewrites, and its probabilities were only moderately calibrated (ECE 0.129). The margin over the perplexity detectors is specific to this setting: in the 30-text development pilot on commercial leaderboard outputs, which has no matched human texts, Binoculars (0.80) and Fast-DetectGPT (0.84) were level with Jev (0.78).

\begin{table}[t]
\centering\small\setlength{\tabcolsep}{4.5pt}
\caption{Detection on the 92 federal passages: human originals versus AI drafts (92) and versus 12B rewrites (368). The CI resamples the 68 source articles, each with all of its passages. TPR uses a threshold at 5\% false positives on the human texts. $^\dagger$RADAR was added after review under a separately frozen protocol.}\label{tab:det}
\begin{tabular}{lrrrr}
\toprule
Detector & AUROC, draft [95\% CI] & AUROC, rewrite [95\% CI] & TPR@5\%FPR, draft & TPR@5\%FPR, rewrite \\
\midrule
Jev (decision model) & 0.943 [0.915, 0.968] & 0.864 [0.827, 0.902] & 0.728 & 0.565 \\
Fast-DetectGPT & 0.567 [0.494, 0.640] & 0.688 [0.623, 0.752] & 0.076 & 0.201 \\
Binoculars & 0.543 [0.468, 0.617] & 0.654 [0.587, 0.720] & 0.054 & 0.082 \\
RoBERTa detector & 0.435 [0.373, 0.496] & 0.661 [0.610, 0.713] & 0.043 & 0.196 \\
\midrule
RADAR$^\dagger$ & 0.717 [0.666, 0.779] & 0.650 [0.599, 0.708] & 0.043 & 0.052 \\
\bottomrule
\end{tabular}

\end{table}

Two further results concern the perplexity-based detectors. First, as ARB reports for human text rewritten by an LLM \citep{perrone2026arb}, they could not separate the human originals from the AI-styled rewrites: AUROC 0.54 for Binoculars and 0.57 for Fast-DetectGPT overall. On NEW2026 they fell below chance, at 0.37 (0.28--0.47) and 0.39 (0.29--0.47), meaning that the rewrites by 2026 models looked more human to them than the human originals did. Second, humanizing raised their AUROC (to 0.65--0.69) but lowered Jev's (from 0.943 to 0.864). The open humanizer therefore works against the decision model and makes text easier for the perplexity detectors to flag, the direction in which plain paraphrasing also raised Fast-DetectGPT's detection rate in \citet{cheng2025adversarial}. Jev's results agreed between the older GoHumanize passages (0.866) and the 2026 passages (0.864). Training-data overlap is therefore unlikely to explain them.

Formatting differed modestly between groups: in NEW2026 the AI drafts had a median of four paragraph breaks against three for the human originals, and the AI texts were 6--30\% longer. After the main results were known we ran a format counterfactual \citep{voznyuk2026artifact}, collapsing all whitespace to single spaces and scoring the texts with Jev again; after review we repeated it for the other four detectors. Jev's AUROC barely changed: 0.864 to 0.854 for rewrites and 0.943 to 0.936 for drafts. Flattening did not remove the below-chance result of the perplexity detectors on NEW2026 but deepened it. For drafts, Binoculars fell from 0.373 to 0.314 (change $-0.114$ to $-0.009$) and Fast-DetectGPT from 0.386 to 0.302 ($-0.132$ to $-0.046$), so line breaks do not explain it. RADAR moved the other way: flattening raised its AUROC on rewrites from 0.650 to 0.717 (change 0.035 to 0.098), so layout affects it as well. RoBERTa changed by less than 0.05.

For the commercial outputs we have no human texts of the same genres, so detection there is descriptive only. Jev's mean $P(\text{ai})$ was 0.93 on the 33 sources and 0.88 on the 363 humanized outputs, ranging from 0.79 to 0.95 by tool. This included 0.86 for the tool that the commercial detectors judged most human-like (mean human probability 0.96).

\section{Limitations}\label{sec:limits}
\begin{itemize}[leftmargin=1.3em,itemsep=1pt,topsep=2pt]
\item \textbf{Fact checker.} Both validations used AI annotators, and they disagreed on 3 of 10 shared items, mainly on how to treat exact unit conversions. A human audit would be stronger evidence. With precision of about 0.84, a minority of the flagged alterations are false; one was a product name the entity tagger took for an organization. The checker compares values, so it misses changes in the relation between a value and what it describes. In one rewrite, ``rest 2--3 days before the event'' became ``take it easy the day before''. This was counted as retained because the value 2 appears elsewhere, in ``a second ride''. The checker also treats a percentage and a plain number with the same value as identical (``30--40\%'' became ``30--50\%'' while ``40 miles'' remained elsewhere). Our measure is therefore the retention of specified number, name and quoted-span units, not the factual accuracy of the whole text.
\item \textbf{Quoted spans.} The checker treats every span in quotation marks alike, so a reworded sample phrasing or term counts as much as a misquoted statement; in the commercial set such spans are 9 of the 15. The verbatim rule is also strict: expanding or contracting a negation (``did not'' to ``didn't'') counts as a change, which suits verbatim quotation. Quoted spans are 15 of 94 units per commercial source set and 18 of 643 units in the federal set.
\item \textbf{Scope.} Matched-control detection covers one genre (U.S. federal prose), one open humanizer and one decision model, and NEW2026 has only 38 passages. The 12B comparison covers the four released files, whose low-bit versions were trained to offset quantization, so it does not measure the effect of bit width alone. The AI drafts are stylized rewrites of human text, not free generations. The TPR@5\%FPR threshold was set on the evaluation texts. RADAR, the only supervised detector trained against paraphrasing, dates from 2023 and was added after review; no commercial classifier or detector trained on current LLM output, such as DeBERTa-ConPara, was tested. On commercial outputs in the development pilot the perplexity detectors matched Jev. We do not know the decision model's training data. The human originals are 2026 agency text and may themselves have had AI assistance; the Federal Reserve speeches are the least likely to. The commercial detector scores are those published by the leaderboard.
\item \textbf{Mechanism.} We do not know why the perplexity-based detectors fall below chance on the 2026 rewrites. The hard prompt pushes text towards flat, hedged, abstract prose; LLM polishing of human text has been reported to blur the human--AI boundary for reward-based scores \citep{liu2025irm}, and ARB frames the effect as content origin and linguistic surface coming apart \citep{perrone2026arb}. Length differences remain a candidate; line breaks are ruled out by the counterfactual.
\end{itemize}

\section{Ethics and reproducibility}
We audit public tools and detectors and do not propose any evasion method. All texts are in the public domain or released under CC BY 4.0, the open model is released under Apache-2.0, and RADAR's weights are licensed for non-commercial use, which covers this audit. The experiments were planned, run and written up with AI coding assistants under the first author's direction, and the checker validation used AI annotators (Section~\ref{sec:judge}). The fact checker, the per-item unit judgments and detector scores, and a script that recomputes Tables~\ref{tab:fid} and~\ref{tab:det} are available at \url{https://github.com/zwang-devv/humanizer-fact-audit}. The protocols and their freeze hashes, the selection and run code, and the run receipts are archived with checksums and available on request. A rerun of the 552 federal texts on an RTX A6000 reproduced every detector score of the main A40 run exactly. The experiments cost about US\$0.65 in total: \$0.58 of GPU time and about \$0.07 of API calls.

\bibliographystyle{unsrtnat}
\bibliography{references}

\appendix
\section{Prompts and questions}\label{app:prompts}
\paragraph{Jev question.} Each text was passed as the state. The choice question was ``Who produced this text?'' with two options. \emph{human}: ``Written by a human author, without an AI language model generating or rewriting it.'' \emph{ai}: ``Generated by an AI language model, or an AI draft rewritten by an AI tool.''

\paragraph{GoHumanize hard-register prompt.} Copied verbatim from the GoHumanize pipeline (commit dd4bd8a1) and used both for the GoHumanize modern split and for NEW2026. System prompt:
\begin{quote}\small
You rewrite passages into the way a large language model writes when asked to produce polished prose. Every fact, name, number and quotation must survive, but nothing about the wording or the sentence structure should. Do not reuse the passage's sentence boundaries: merge short sentences and split long ones. Do not add headings, lists, commentary or a preamble. Return only the rewritten passage.
\end{quote}
The user message is one of four style instructions, followed by the rules below and then the passage. The four styles are \emph{formal\_polished} (``Rewrite in a polished, formal register: expand contractions, prefer general and abstract vocabulary over concrete or unusual words, smooth out irregular sentence rhythms so sentences are of similar length, and open sentences with connective phrases such as `Additionally', `Furthermore', `Moreover', `In essence', `It is worth noting that'.''), \emph{explanatory} (``Rewrite as a clear, neutral explanation: state things plainly, add light framing such as `This highlights', `This suggests', `Ultimately', remove idiosyncratic phrasing and dialect, and keep a calm, even, slightly impersonal tone throughout.''), \emph{hedged\_helpful} (``Rewrite in a careful, helpful assistant tone: soften strong statements with hedges (`may', `can often', `tends to'), use balanced sentence structures, replace vivid or archaic words with common modern equivalents, and close with a tidy summarising sentence.'') and \emph{corporate\_clean} (``Rewrite in clean, efficient modern prose as a professional writing assistant would produce: short paragraphs, consistent sentence length, generic descriptive adjectives, transitional words between sentences, no contractions, no slang, no fragments.''). Rules:
\begin{quote}\small
Apply these as well:
\begin{itemize}[leftmargin=1.3em,itemsep=0pt,topsep=1pt]
\item Open most sentences with a connective or framing phrase (Additionally, Furthermore, Notably, It is worth noting that, In essence, Ultimately).
\item Turn verbs into noun phrases where you can: `decided' becomes `reached a decision', `fell' becomes `experienced a decline'.
\item Replace concrete or idiomatic wording with general, abstract equivalents.
\item Hedge claims with `may', `can often', `tends to', `appears to'.
\item Make every sentence a similar length, roughly 22 to 30 words, so the rhythm is flat.
\item Add framing and summarising clauses so the passage runs 20 to 40 percent longer than the original, and close with a tidy summarising sentence.
\item Use no contractions, no fragments, no rhetorical questions, no dashes.
\end{itemize}
\end{quote}
In NEW2026 the style was assigned by passage rank so that each generator covered all four styles; GoHumanize assigns styles the same way. Both generators are reasoning models and were run with low reasoning effort; Gemini used the pipeline's temperature of 0.8, and GPT-6-Luna, which does not accept a temperature, used its default.

\section{Unit alteration without quoted spans}\label{app:quotefree}
Table~\ref{tab:quotefree} repeats the document-level alteration rate of Table~\ref{tab:fid} after removing quoted-span units, so that a document counts as altered only if a number or name unit changed. The denominator is the number of eligible documents that contain at least one number or name unit.

\begin{table}[!htbp]
\centering\small
\caption{Documents with at least one altered unit, with and without quoted-span units. Commercial tools: AIHB October 2026. 12B: the open humanizer on the federal drafts.}\label{tab:quotefree}
\begin{tabular}{lrrrr}
\toprule
Tool & Eligible docs & With quoted spans & Altered, all units & Altered, numbers and names only \\
\midrule
aihumanize & 20 & 6 & 13/20 (0.65) & 9/19 (0.47) \\
writehuman & 20 & 6 & 13/20 (0.65) & 11/19 (0.58) \\
stealthgpt & 20 & 6 & 11/20 (0.55) & 10/19 (0.53) \\
undetectable-ai & 20 & 6 & 11/20 (0.55) & 6/19 (0.32) \\
cleverhumanizer & 20 & 6 & 10/20 (0.50) & 7/19 (0.37) \\
gptinf & 20 & 6 & 9/20 (0.45) & 5/19 (0.26) \\
rehumanize & 20 & 6 & 9/20 (0.45) & 4/19 (0.21) \\
smarthumanizer & 20 & 6 & 8/20 (0.40) & 5/19 (0.26) \\
hix-bypass & 20 & 6 & 7/20 (0.35) & 2/19 (0.11) \\
superhumanizer & 20 & 6 & 7/20 (0.35) & 4/19 (0.21) \\
undetectedgpt & 20 & 6 & 7/20 (0.35) & 4/19 (0.21) \\
\midrule
12B bf16 & 90 & 13 & 11/90 (0.12) & 5/89 (0.06) \\
12B Q8\_0 & 90 & 13 & 15/90 (0.17) & 7/89 (0.08) \\
12B Q3-QAT & 90 & 13 & 16/90 (0.18) & 10/89 (0.11) \\
12B IQ2\_XS-QAT & 90 & 13 & 17/90 (0.19) & 9/89 (0.10) \\
\bottomrule
\end{tabular}

\end{table}
\FloatBarrier

\section{Article-level versus passage-level intervals}\label{app:cluster}
The 92 federal passages come from 68 articles: the 54 GoHumanize passages from 30 articles (up to four from one article) and the 38 NEW2026 passages from 38. The frozen analyses resampled passages. Table~\ref{tab:cluster} compares those intervals with intervals that resample articles, each with all of its passages and their versions (2,000 replicates, the same seed). All federal-set intervals in the main text, Table~\ref{tab:det} and Figure~\ref{fig:det} resample articles. NEW2026 has one passage per article, so for it the two schemes coincide.

\begin{table}[!htbp]
\centering\small
\caption{Federal-set 95\% intervals under the frozen passage resampling and under article resampling. $^\dagger$Post hoc (RADAR and the whitespace counterfactual).}\label{tab:cluster}
\begin{tabular}{lrrr}
\toprule
Quantity & Estimate & Passages (frozen) & Articles \\
\midrule
Jev AUROC, human vs.\ rewrite (H3) & 0.864 & [0.826, 0.902] & [0.827, 0.902] \\
Jev $-$ Binoculars, rewrite & 0.211 & [0.138, 0.284] & [0.132, 0.294] \\
Jev $-$ Fast-DetectGPT, rewrite & 0.176 & [0.107, 0.248] & [0.099, 0.256] \\
Jev $-$ RADAR$^\dagger$, rewrite & 0.215 & [0.166, 0.263] & [0.157, 0.264] \\
RADAR$^\dagger$, flattened $-$ original, rewrite & 0.067 & [0.037, 0.098] & [0.035, 0.098] \\
Retention, bf16 $-$ IQ2\_XS-QAT (H2b) & 0.008 & [-0.004, 0.021] & [-0.004, 0.020] \\
Evasion, bf16 $-$ IQ2\_XS-QAT (H2b) & -0.003 & [-0.065, 0.046] & [-0.065, 0.048] \\
\bottomrule
\end{tabular}

\end{table}
\end{document}